\documentclass[10pt,twocolumn]{article} 
\usepackage{simpleConference}
\usepackage{times}
\usepackage{graphicx}
\usepackage{amssymb}
\usepackage{url,hyperref}

\usepackage{microtype}
\usepackage{graphicx}
\usepackage{booktabs}

\usepackage{amsmath}
\usepackage{mathtools}
\usepackage{amsthm}

\usepackage{natbib}
\renewcommand{\cite}[1]{\citep{#1}}

\usepackage[capitalize,noabbrev]{cleveref}

\usepackage{indentfirst}
\usepackage{color}
\usepackage{bm}
\usepackage{soul}
\usepackage{amsmath}
\usepackage{amssymb}
\usepackage{amsthm}
\usepackage{lipsum}
\usepackage{multirow}
\usepackage{booktabs}
\usepackage[labelformat=simple]{subfig}

\newcommand{\x}{\bm{x}}
\newcommand{\w}{\bm{w}}
\newcommand{\y}{\bm{y}}
\newcommand{\ep}{\mathbb{E}}

\DeclareMathOperator*{\argmax}{arg\,max}

\theoremstyle{plain}

\newtheorem{theorem}{Theorem}[section]

\newtheorem{proposition}[theorem]{Proposition}

\theoremstyle{definition}

\theoremstyle{remark}

\usepackage[textsize=tiny]{todonotes}

\begin{document}

\title{A Unified Study of Machine Learning Explanation Evaluation Metrics}

\author{Yipei Wang, Xiaoqian Wang\thanks{corresponding author}\\
Elmore School of Electrical and Computer Engineering \\
Purdue University \\
West Lafayette, IN 47907 \\
\texttt{wang4865@purdue.edu, joywang@purdue.edu} \\
}

\maketitle
\thispagestyle{empty}

\begin{abstract}
The growing need for trustworthy machine learning has led to the blossom of interpretability research. Numerous explanation methods have been developed to serve this purpose. However, these methods are deficiently and inappropriately evaluated. Many existing metrics for explanations are introduced by researchers as by-products of their proposed explanation techniques to demonstrate the advantages of their methods. Although widely used, they are more or less accused of problems. We claim that the lack of acknowledged and justified metrics results in chaos in benchmarking these explanation methods -- \emph{Do we really have good/bad explanation when a metric gives a high/low score?} We split existing metrics into two categories and demonstrate that they are insufficient to properly evaluate explanations for multiple reasons.
We propose guidelines in dealing with the problems in evaluating machine learning explanation and encourage researchers to carefully deal with these problems when developing explanation techniques and metrics. 
\end{abstract}

\section{Introduction}
The rapid development of deep neural networks (DNNs) has achieved great success, leading to widespread applications in reality. However, the applications of such black boxes in high-stake areas have drawn more and more concerns. General Data Protection Regulation (GDPR) stipulates a right to explain algorithmic decision making. 
Due to the growing needs, research in Explainable Artificial Intelligence (XAI) -- especially explainable machine learning, has been a hit in recent years. 
As a consequence, a rich genre of explanation methods have been proposed, serving the purposes of exploring and studying the inner mechanism of black-box machine learning models.

Despite the prosperity in explanation methods, recent studies show that a nonnegligible portion of such explanation methods have severe intrinsic flaws~\cite{adebayo2018sanity, srinivas2020rethinking, shah2021input}. Besides, even established on reasonable criteria, different explanation methods provide different explanations for the same model-input pair, which can be confusing and undermine the trust from end users, because it is hard to decide \textit{which} explanation to trust if they are contradictory.
This chaos makes finding the proper explanation method as difficult as understanding a black-box model for end users -- The users have to actually know the mechanism of the black-box model to find the desired explanation method, which is a vicious circle.
Therefore, metrics that are used to evaluate explanation methods have come into play.

\begin{figure}[t!]
    \centering
    \includegraphics[width = 1\columnwidth]{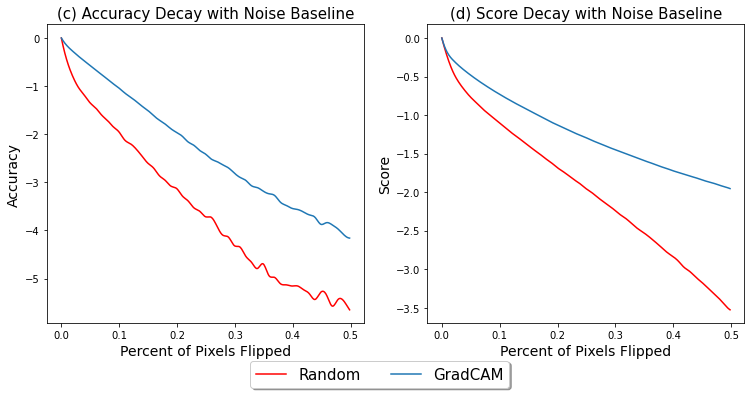}
    \caption{Results of using Pixel Flipping metric~\cite{samek2016evaluating} to evaluate two explanations (GradCAM~\cite{selvaraju2017grad} explanation and Random explanation) of VGG-16 model on ImageNet. Pixels are masked in descending order of attribution values. Random masking yields faster decay than GradCAM, which according to Pixel Flipping metric, shows that Random explanation captures ``more important'' pixels than GradCAM in explanation.}
    \vspace{-2pt}
    \label{fig:first_example}
\end{figure}

Based on the difference in measured statistics, quantitative metrics for explanations can be roughly divided into two categories. The first category is the \textit{alignment} metric. One can check if the explanations correspond to the prior knowledge.
The other category is the \textit{performance} metric, where the input samples are usually modified with respect to the explanations in some manners and then fed to the prediction model. The change of performance of the prediction model is used to evaluate the corresponding explanation method.
These two categories of metrics have been widely used by researchers in both developing explanation methods and metrics.
{However, even though some problems of evaluating explanation methods with the different metrics have been noticed elsewhere, they have never been paid sufficient attention to.}
The lack of in-depth study on issues with using existing explanation metrics can be detrimental, and lead to chaotic situations in the entire XAI area.

Here we take one of the most widely used performance metrics, Pixel Flipping~\cite{samek2016evaluating}, as an example. In this metric, we mask pixels of input images according to attribution values based on an explanation method and then test performance decay as more pixels are masked. A good explanation method will ideally result in faster performance decay when pixels are masked in the descending order of attribution values. However, as shown in \cref{fig:first_example}, Random attribution (which assigns attribution values randomly) has faster decay than GradCAM~\cite{selvaraju2017grad} in terms of both the log ratio of accuracy and predicted scores. But this by no means implies that Random attribution is a better explanation method than Grad-CAM. As a result, in this paper, we propose to ask the question, \textit{``Do we really have a good/bad explanation when an evaluation metric says so?''}. In order to answer this question, we explore the reasons behind this chaos and carry out a unified study on the evaluation metrics for explanations.

Our contribution can be summarized as follows:
\vspace{-7pt}
\begin{itemize}
    \setlength\itemsep{0em}
    \item We categorize existing metrics for attribution methods into \textit{alignment} metrics and \textit{performance} metrics and carry out various experiments in case studies to demonstrate the intrinsic flaws of both categories.
    
    \item We demonstrate that incompatible goals between explanation methods and metrics lead to unfair evaluation.
    
    \item We use \textit{faithfulness} and \textit{plausibility} to analyze the reasons behind the chaos of explanation evaluation. We introduce \textit{projected gradient descent (PGD) enhancement} to illustrate the conflict between such two aspects.
    
    \item We propose criteria that guide how to avoid being misled by the chaos of explanation evaluation metrics.
\end{itemize}

\section{Related Work}
\label{sec:related_work}
{In this section we review related works that develop or evaluate explanation/interpretation in machine learning. We clarify that the difference between notions \textit{explainability} and \textit{interpretatability} is beyond the scope of this work. They are not distinguished and will be used interchangeably in the following context.}

\textbf{Explanation Methods} 
Depending on the ways explanations are expressed, explanation methods can be divided into attribution methods and high-level methods. Given input data, attribution methods assign each feature (pixel in images, token in text, etc.) of the input a value, representing the importance or relevance of the feature to the output.
{Based on the mechanism of producing attribution explanations, there are back-propagation methods~\cite{simonyan2013deep,zeiler2014visualizing,springenberg2014striving,bach2015pixel,zhou2016learning,selvaraju2017grad,sundararajan2017axiomatic,shrikumar2017learning,zhang2018top,parekh2020framework}, etc. and perturbation methods~\cite{petsiuk2018rise,lundberg2017unified,fong2019understanding}, etc. Differently, high-level methods provide explanations for the prediction process from higher levels, such as concept-based explanations~\cite{ribeiro2016should, kim2018interpretability,zhou2018interpretable, ghorbani2019towards, koh2020concept}, {sample-based explanations}~\cite{koh2017understanding, chen2020oodanalyzer}, etc.
Due to the space limit, we refer to \cref{app:related_work} for other ways of characterizing explanation methods.

Among the different forms of explanation methods, attribution methods are more universal and have always been evaluated quantitatively together using the same metric \cite{petsiuk2018rise,fong2019understanding}. While for other forms, a unified comparison can be difficult. For example, different concept-based methods may have different sets of concepts or even be developed for different datasets/tasks. Thus in this paper, we focus on metrics for attribution methods.
}

\textbf{Metrics for Evaluating Attribution Methods}
Depending on the statistics measured by the metrics, attribution metrics can be divided into \textit{alignment} metrics and \textit{performance} metrics. Alignment metrics measure how well the explanation aligns with prior knowledge or given supervision information. \citet{selvaraju2017grad} use weakly-supervised localization to evaluate the explanations. Following a similar idea, \citet{zhang2018top} simplify it by introducing Pointing Game, which calculates the ratio of the number of samples whose attribution maps can correctly point at the pre-annotated object area. \citet{poerner2018evaluating, yang2019benchmarking, adebayo2020debugging, zhou2021feature} forge new datasets trying to introduce artificial ground truth.

In contrast, performance metrics measure the change in prediction model performance w.r.t. certain modification in input samples. \citet{samek2016evaluating} introduce Pixel Flipping, which calculates the performance change when input features are perturbed based on the explanations. Most performance metrics share the same idea as Pixel Flipping and resemble it. They modify the input according to the explanations and observe the change in predictions, such as the insertion/deletion metric~\cite{petsiuk2018rise}, masking top pixels~\cite{chen2018shapley}, top-$k$ ablation~\cite{sturmfels2020visualizing}, Impact Score~\cite{lin2019explanations}, word deletion~\cite{arras2016explaining}, infidelity~\cite{yeh2019fidelity}, etc. \citet{hooker2018benchmark} introduce ROAR, which is a modification of Pixel Flipping and requires training a completely new black-box model every time the number of masked pixels changes.

\textbf{Guidelines}
\citet{preece2018stakeholders} argue that different stakeholder communities have different expectations of machine learning explanations. \citet{miller2019explanation} carry out an analysis from the social scientific perspective. \citet{jacovi2020towards} propose guidelines in evaluating explanations according to faithfulness and plausibility. \citet{tomsett2020sanity} perform sanity checks on saliency metrics, {which studies the properties of metrics by introducing the reliability from the psychometric testing.}

\section{Incompatible Goals}
\label{sec:goals}
{Denote by $f:\mathbb{R}^d\rightarrow\mathbb{R}^c$ a prediction model before the last activation, where $d$ is the dimension of input data, $c$ is the dimension of the output. Then the attribution method for $f$ is defined as $\phi_f^i:\mathbb{R}^d\rightarrow\mathbb{R}^d$. (For RGB images the codomain of $\phi$ should be $\mathbb{R}^{d/\mathrm{channels}}$, but here we omit this difference.) $i\in [c]$ corresponds to the $i$-th class, and we will use $t$ when referring to the true class. Since all attribution methods share the same form, it may seem natural to evaluate them altogether, with some metric $M$. However, we argue that different attribution methods may focus on different goals, and hence should not always be evaluated together.}

{
{For example, despite it is preferred by many researchers that an explanation method should satisfy $\phi_f^i(\x) = f(\x)_i,\forall i\in [c]$~\cite{bach2015pixel, sundararajan2017axiomatic, lundberg2017unified}, which we refer to as \textit{completeness},}
a lot of attribution methods do not satisfy this property. But \textit{should they?}} The Gradient method does not satisfy completeness, but it does not focus on reassigning $f(\x)_i$ to the input in the first place. For methods where attribution values are \textit{additive contributions} to the predictions, this is a reasonable property, but the attribution values from the gradient method only represent the \textit{sensitivity} of the prediction $f(\x)_i$ w.r.t. the corresponding input features in their neighborhoods. Also, suppose $f$ has only one layer and degenerates to a linear model such that $f(\x) = W^T\x$, where $W\in\mathbb{R}^{d\times c}$. Then $\forall i\in[c]$, {if we set $w^{(i)}_jx_j$ as the attribution value for $x_j$ as the explanation,} it satisfies completeness. However, the gradient $\nabla_{\x} f(\x)_i = \w^{(i)}$ does not. But this explanation is as important as the complete explanation $w^{(i)}_jx_j$. Therefore, one has to be aware of the goals of both the explanation methods and the metrics. A metric can reasonably evaluate an explanation only if their goals are compatible.

\section{The Pitfall of Plausibility}
\label{sec:faith_ration}
There are two main bodies in XAI research, machines and humans. The goal of XAI is to convey information from machines to humans, in a comprehensible way.
Machine wise, the \textit{faithfulness} of explanation refers to that explanation reflects the true mechanism of the prediction model $f$. While human wise, \textit{plausibility} refers to that explanation conforms with humans' perception.
It may seem natural to expect explanations that are both faithful and plausible. However, these two aspects can be contradictory and easily confused in the development and evaluation of explanations.

\begin{table}[!t]
\centering
\caption{Model-Explanation-Metric triple combinations. The bottom row shows metric evaluation results, provided the metric measures \textit{plausibility}
of explanation. ``$\checkmark$'' indicates high metric score, and ``$\times$'' indicates low score.}
\small
\begin{tabular}{c|c|c|c|c}
\toprule
Model $f$ & \multicolumn{2}{c|}{Plausible} & \multicolumn{2}{c}{Implausible}\\
\hline
Explanation $\phi$ & {Faithful} & {Unfaithful} & {Faithful} & {Unfaithful}  \\
\midrule
Metric $M$ & $\checkmark$ & $\times$ & $\times$ & Uncertain\\
\bottomrule
\end{tabular}
\label{tab:model_explanation_metric}
\end{table}

Many existing explanation metrics -- including all alignment metrics, lean towards plausibility. They evaluate explanations by checking if the explanations are compatible with some prior knowledge. Such prior knowledge is \textit{independent} of model $f$.
This is a tempting yet dangerous way, because the true mechanism of model $f$ is not necessarily plausible, and may not correspond to the prior knowledge humans possess.

On the one hand, a model $f$ can make predictions in a way that is completely different from human knowledge, such as textures~\cite{geirhos2018imagenet}, controlled ground truth~\cite{kim2018interpretability}, overinterpretations~\cite{carter2020overinterpretation}, perturbations by the adversarial attack, overfitting features, etc. Although they are all recognized to have a great influence on the mechanism of the black-box models, none of them are plausible to humans, and none of them can be easily included in the prior knowledge. On the other hand, a plausible ``explanation'' can have nothing to do with the predictive mechanism of the black-box model, such as edge detectors~\cite{adebayo2018sanity}. 

Therefore, metrics that focus on plausibility will not provide useful information about the inner mechanism. Even worse, it can jeopardize the evaluation of explanations. As shown in \cref{tab:model_explanation_metric}, when the mechanism of the model itself is plausible (e.g. focusing on objects when classifying), a \textit{faithful} explanation will also be plausible. Hence this flaw is not very evident. However, when the model is implausible, a \textit{faithful} explanation will capture the implausibility and is thereby implausible. In this case, it will be unfairly evaluated by plausibility metrics.

\section{Projected Gradient Descent Enhancement}
In order to demonstrate how plausibility can undermine faithfulness, here we propose {projected gradient descent (PGD) enhancement, which is the inverse of PGD attack~\cite{madry2017towards}}. PGD enhancement uses exactly the same criterion as PGD attack, but changes the direction of each step. As a result, given a randomly initialized, untrained neural network, it produces enhanced samples with human-imperceptible perturbations. Such enhanced samples can be correctly classified by the untrained model. We test AlexNet~\cite{krizhevsky2014one}, VGG-16~\cite{simonyan2014very}, and MnasNet~\cite{tan2019mnasnet} on the validation set of ISLVRC 2012~\cite{deng2009imagenet}. Please refer to \cref{app:PGD_enhancement} for implementation details.

As shown in \cref{tab:PGD_enhancement}, given an enhanced validation set, an untrained model can have comparable performance to a well-trained model (trained and tested on the raw datasets). 
Besides, the desired explanation method should give them very different explanations since they have intrinsically different mechanisms (trained and untrained).

\begin{figure}[!t]
\centering
\includegraphics[width = \linewidth]{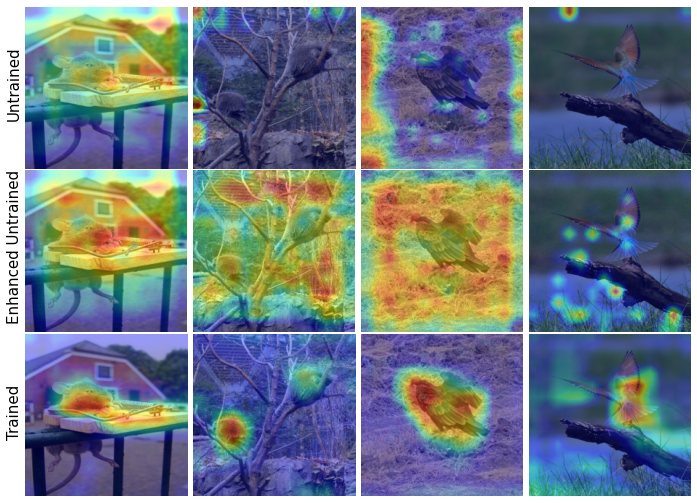}
\caption{Grad-CAM explanations of VGG-16 models. The results are from the untrained model with raw data as input (top row), the untrained model with PGD-enhanced data as input (middle row), and the trained model with raw data as input (bottom row). All samples of the middle and the bottom rows are correctly classified, and all samples from the top row are falsely classified.}
\label{fig:PGD_enhancement}
\end{figure}

We take VGG-16 and Grad-CAM as an example and plot the explanation heatmaps in \cref{fig:PGD_enhancement}. Please refer to \cref{app:PGD_enhancement} for other explanation methods.
In \cref{fig:PGD_enhancement}, there is no doubt that explanations in the bottom row are the most plausible ones among rows. The objects are highlighted precisely, and intuitively, very little uninformative area is highlighted. However, if evaluated in the sense of plausible explanation, one will justify that Grad-CAM is bad based on the second row. This judgement is neither fair nor reasonable since it may correspond to the third case in \cref{tab:model_explanation_metric}, where the model itself is implausible while the explanation is faithful to this implausibility. Then a plausibility-based metric will unfairly evaluate the explanation.

This example demonstrates the reason why plausibility should be used carefully in metrics to evaluate explanations. Explanations should first be faithful and can reflect the true mechanism of the model. {In \cref{fig:PGD_enhancement}, the differences in mechanism between the middle and the bottom row should be ascribed to two factors: 1) the differences between the trained and the untrained VGG-16 networks; and 2) the human-imperceptible PGD enhancement. However, neither the level of training in the model nor the enhancement in data is plausible to humans.} As a consequence, we suggest that plausibility should be specifically separated from faithfulness in evaluations, and both aspects should be carefully considered in {both developing explanation methods and evaluation metrics.} Otherwise, the chaos caused by the confusion of faithfulness and plausibility will impede the interpretability research. We further look into the chaos in explanation evaluation in the following case studies.

\begin{table}[!t]
\centering
\caption{Accuracy comparison between the trained models on raw data and the untrained models on PGD-enhanced data. Three different model structures (AlexNet, VGG-16, MnasNet) are used.}
\begin{tabular}{cc|cc}
\toprule
        &           & Raw      & Enhanced \\
\midrule
\multirow{2}{*}{AlexNet}   & Untrained & 0.086\%  & \textbf{64.664\%} \\
& Trained   & 52.558\% & -        \\ \hline
\multirow{2}{*}{VGG-16} & Untrained & 0.102\%  & \textbf{99.956\%}         \\
& Trained   & 72.230\% & -\\ \hline
\multirow{2}{*}{MnasNet} & Untrained & 0.110\%  & 64.588\%         \\
& Trained   & \textbf{71.718\%} & -\\
\bottomrule
\end{tabular}
\label{tab:PGD_enhancement}
\end{table}

\begin{figure*}[!t]
    \centering
    \includegraphics[width = 1.0\textwidth]{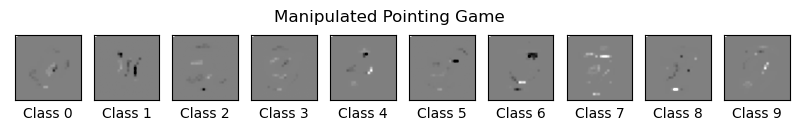}
    \caption{The explanation $\w_i\odot\x$ of the manipulated sparse linear model for a digit 7 from MNIST dataset. We constrain the sparse linear model such that the top left corner pixel has the highest attribution score. The explanation from such model is faithful but has very low Pointing Game ratio.}
    \label{fig:PointingGame_counterexample}
\end{figure*}

\section{Case Studies}
\label{sec:case_studies}
Among various explanation metrics, we take one representative and widely used metric in each category for the case study and demonstrate the chaos. In the demonstration, we use sparse linear models with Lasso regularization as the benchmark for \textit{faithfulness} in all experiments. The reasons for choosing such a benchmark include two folds: 1) sparse linear model is an attribution method; 2) white-box models like sparse linear models can axiomatically reflect the true mechanism of itself while building a truly faithful post-hoc explanation for DNNs still remains an open question.
We use the Hadamard product $\hat{\w}^{(i)}=\w^{(i)}\odot\x\in\mathbb{R}^d,i\in[c]$ as the attribution explanation of the sparse linear model.

\subsection{Alignment Metrics}
Among alignment metrics that measure how well explanations conform with prior knowledge, we take Pointing Game~\cite{zhang2018top} as a representative.
Pointing Game is one of the most popular metrics that measure attribution explanation methods. Please refer to \cref{app:pointing_game} for more details of this metric. Here, the annotated bounding boxes (or the object segmentation) are the prior knowledge. This prior knowledge will directly result in a plausibility metric since the annotations are solely based on the way we humans perceive the world -- the objects. Intuitively, pixels that contribute highly to the prediction of the targeted classes should be on the objects, or at least be in the corresponding bounding boxes. However, if a prediction of a model $f$ is based on human-imperceptible features, (which, as discussed earlier, is not a rare scenario) then Pointing Game fails to properly evaluate the correctness of an explanation $\phi_f$. Besides, if we have $\phi_f \equiv \phi$ independent from $f$ to be an edge detector, it will gain a high score in Pointing Game, but has no meaning in explaining the model $f$.

Here we take the sparse linear model as an example since it is axiomatically acknowledged as faithfully interpretable. However, such a sparse linear model can be too simple for complex image datasets with annotations, such as Pascal VOC~\cite{everingham2010pascal}, COCO~\cite{lin2014microsoft}, etc. Therefore, we build an annotated MNIST dataset with bounding boxes.
Please refer to \cref{app:MNIST} for the illustration of bounding boxes for some examples.
It is expected by Pointing Game that the pixel with the highest attribution value should be in the bounded area. This is very intuitive and tempting. However, it can be easily demonstrated that this metric is not consistent with the true mechanism of $f$. For sparse linear models, the true (additive) mechanism is the Hadamard product $\hat{\w}^{(t)}\odot \x$. We modify the weights corresponding to the top left corner of all classes, so that the top left pixel has the largest attribution values in all classes. 
In this way, on the one hand, the most contributing pixel in Pointing Game will always point to the top left corner. On the other hand, since the manipulation raises the prediction scores of all classes by the same amount, the classification results stay the same.
Besides, here only one weight is changed from zero to non-zero, and the sparsity is preserved, so the model is still a sparse linear model, and it is still recognized as a self-interpretable model.

We show an example of explanations of input digit $7$ in \cref{fig:PointingGame_counterexample}. It can be found that the left corner has the highest attribution value in the classification process, but does not conform with the prior knowledge in Pointing Game. The comparison results are shown in \cref{tab:Pointing_game}. We can find that Pointing Game gives very delusive results. Through such minor and imperceptible change, the sparse linear model preserves all properties including the expressiveness (same accuracy), but the Pointing Game score drops from 100\% to 0. That is, a human-imperceptible change in model explanation can make a drastic change in the Pointing Game ratio, which means it is also very sensitive.

As a representative of alignment metrics, the above analysis on Pointing Game can be easily generalized to other metrics of this category.
We argue that the mismatch between the explanations and the prior knowledge should be ascribed to the model $f$, instead of the explanation $\phi_f$. On the contrary, the obligation of explanations is indeed to find the \textit{mismatch} between $f$ and prior knowledge. In this way, the explanations enable us to debug the model $f$, and build more transferable models.

\begin{table}[!t]
\vspace{-10pt}
\caption{Pointing Game ratio results.}
\centering
\small
\begin{tabular}{c|c|c}
\toprule
& Standard Sparse & {Manipulated Sparse}\\
& Linear Model & Linear Model\\
\midrule
Accuracy  & 79.51\%                         & 79.51\%                \\
\hline
Pointing Game Ratio & {100.0\%  } & {0.0\%  } \\
\bottomrule
\end{tabular}
\label{tab:Pointing_game}
\end{table}

\subsection{Performance Metrics}

Among performance metrics that measure changes in model performance w.r.t. certain modifications in input samples, Pixel Flipping~\cite{samek2016evaluating} is a typical representative in this category. Other members such as word deletion~\cite{arras2016explaining}, masking top pixels~\cite{chen2018shapley}, the deletion/insertion metric~\cite{petsiuk2018rise}, top-$k$ ablation~\cite{sturmfels2020visualizing}, etc. can be seen as variations of Pixel Flipping. All of these variations are based on the same criteria, that is, to mask (or insert for the insertion metrics) pixels by the order of their importance scores assigned by the explanation method. Usually, the change of prediction score $f(\x)_t$ (and change of accuracy in other versions) are measured with respect to the portion of masked pixels in image data (or other kinds of features in other data types). For clarity, we will focus on Pixel Flipping in the following context. It is also worth emphasizing that the flaws of Pixel Flipping are shared by other performance metrics.

Given an attribution method $\phi_f^t(\x)$, all $d$ features can be ranked by an index set $A = \{\alpha_1,\alpha_2,\cdots,\alpha_{d}\}$, which is a permutation of $[d]$. Pixel Flipping masks pixels by the order of the index set $A$. That is, $x_{\alpha_1}, x_{\alpha_2}, \cdots, x_{\alpha_d}$. Generally, the index set $A$ is built by the monotonically decreasing order, such that $\phi_f^t(\x)_{\alpha_1}\ge \phi_f^t(\x)_{\alpha_2}\ge \cdots \phi_f^t(\x)_{\alpha_{d}}$,
where a more important pixel decided by the explanation method will be masked earlier. Therefore, when evaluating $\phi_f$, a faster decay on the prediction score (or the accuracy) is more desired. It can be noticed that Pixel Flipping aims at capturing the inner mechanism of $f$. It achieves this by making the decoupling assumptions on input features. This is a very strong assumption and is only true for linear additive models. As a consequence, the flaws of Pixel Flipping are from multiple aspects.

\subsubsection{The Indicator Flaw}
Due to the independence assumption of Pixel Flipping, we know that if $f$ is a linear model, then the assumption holds, and that in this case Pixel Flipping is faithful. However, we argue that the indicator should be the change in predicted score instead of accuracy for Pixel Flipping. If the accuracy is used as the indicator, Pixel Flipping fails even for linear models. This is actually from the formulation of attribution methods. No matter based on backpropagation or perturbation, attribution methods are always subject to some selected target $f(\x)_i, i\in[c]$, which does not directly affects the accuracy. The accuracy is determined by the final predicted result $t' = \argmax_{i\in [c]}f(\x)_i$, which contains the information of all classes. Now that $\phi_f^t$ itself does not contain information about the predicted results of classes $\forall j, j\ne t$, using accuracy as the indicator is an ill-posed way to evaluate $\phi_f$. It is not guaranteed that masking the pixel with the truly highest contribution will result in the fastest decay in accuracy, because an input feature being the \textit{most important} to the true class $t$ does not mean that it is not as important to the class $j$. Please refer to \cref{app:appendix_accuracyPixelFlipping} for the formal statement and the construction of the counterexample. 

\begin{figure*}[htb!]
\centering
\includegraphics[width = 1.0\textwidth]{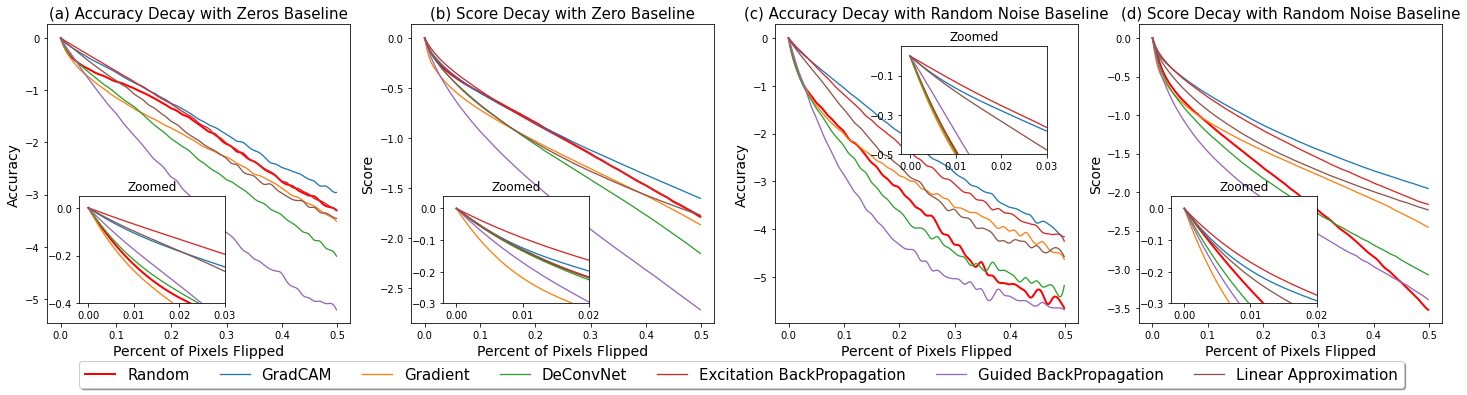}
\caption{Pixel Flipping results with the zeros baseline (left two subfigures) and the random noise baseline (right two subfigures). All results are normalized to $[0,1]$ and then taken the logarithm for better illustration.}
\label{fig:PixelFlipping_results}
\end{figure*}

\subsubsection{The Goals Flaw}
As mentioned in \cref{sec:goals}, the goal of the metric $M$ has to be compatible with the goal of the attribution method $\phi_f$ to properly evaluate the attribution method. For Pixel Flipping, the goal is similar to completeness. {If $\phi_f$ represents sensitivity, relevance, or basically any property other than the additive contribution, it is not appropriate to evaluate them with Pixel Flipping.}
We still take $f$ to be linear as an example, since DNNs will introduce too much exogenous bias as we explain later. Let $c = 1, f(\x) = \w^T\x$. If we use the Hadamard product as the explanation, where the attribution value for $x_j$ is $w_jx_j$, then $\phi_f$ has the same goal as Pixel Flipping -- to assign the additive contribution. However, if we use the weight $w_j$ as the attribution value for $x_j$, its goal is no longer the same as Pixel Flipping. The attribution value $w_j$ indicates how \textit{changing} $x_j$ influences the prediction. And correspondingly, the Hadamard product explanation aces Pixel Flipping, because if the attribution value $w_jx_j$ is largest among unmasked values, then masking $x_j$ results in the fastest decay in the performance. However, in the weight explanation, masking $x_j$ with the largest attribution value $w_j$ does not guarantee the largest decay. This demonstrates the goals flaw of Pixel Flipping.

\subsubsection{The Distribution Flaw}
One of the most fatal flaws of Pixel Flipping is the assumption of distributions. It implicitly assumes the independence among input features, which can be a too strong assumption for complex datasets or complex models. It is also argued that since $f$ is trained on the original dataset, it is unreasonable to test its performance on the masked dataset since they have different distributions~\cite{hooker2018benchmark}.

We quantitatively demonstrate the out-of-distribution flaw of Pixel Flipping by introducing a random noise case in the comparisons among attribution methods. The experiments are carried out on the ILSVRC 2012 validation dataset. We use Pixel Flipping to measure several popular explanation methods\footnote{{We clarify that due to the goals flaw, we disagree that some of these methods should be evaluated by performance metrics. The reason we include these methods in the evaluation here is not to show which explanation technique is better, but to illustrate the pitfalls of Pixel Flipping as an evaluation metric.}}, including Grad-CAM~\cite{selvaraju2017grad}, Gradient~\cite{simonyan2013deep}, DeConvNet~\cite{zeiler2014visualizing}, Excitation Back-propagation~\cite{zhang2018top}, Guided Back-propagation~\cite{springenberg2014striving}, and Linear Approximation. Also, we include the random masking criterion as the benchmark, where the pixels are masked at uniform random.
Images are resized to $224\times 224$, and in the illustration, $9k (\approx 18\%)$ pixels are masked for each sample.
The measures are the predicted scores $f(\x)_t$.

According to Pixel Flipping, the faster $f(\x)_t$ decays, the better $\phi_f$ is. It is intuitively desired by anyone that random masking will have the worst performance, i.e., the slowest decay. However, as shown in \cref{fig:PixelFlipping_results}, we can find the counterintuitive results. At the beginning, random masking has a quite fast decay compared with explanation-based masking. As the portion of masked pixels increases, this phenomenon gradually diminishes.

Another very interesting observation from the experiment results in \cref{fig:PixelFlipping_results} is that, if we pay attention to the explanation methods being tested, we can find that Guided BP, DeConvNet, and Gradient always outperform others. These three methods share a very important similarity with random masking -- Their masked pixels are isolated, while the other three methods' masked pixels always adjoin. {Please refer to \cref{app:PGD_enhancement} for some illustrations.} In addition to the ``informative'' pixels themselves, the distribution difference among explanation methods also has great impact on performance decay.

\begin{figure}[!t]
    \centering
    \includegraphics[width = 0.45\textwidth]{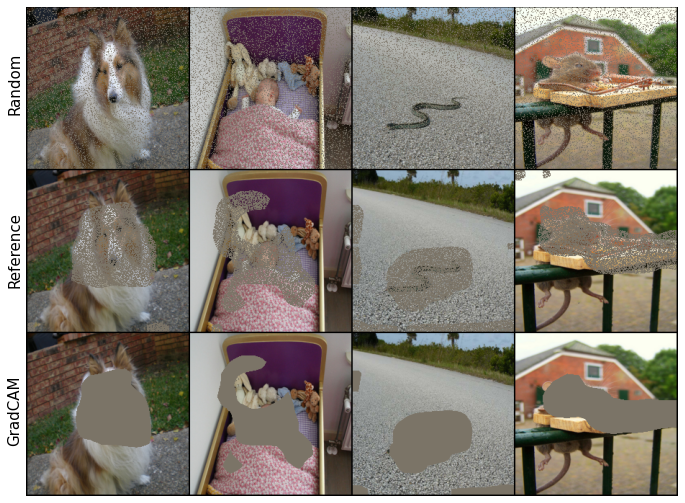}
    \caption{Illustrations of masked images in Pixel Flipping of VGG-16. For each sample, $n=$ 9k pixels are masked. The first row is random masking. The second row is random masking with Grad-CAM reference (i.e., the masked 9k pixels are randomly chosen from the $N$=12k pixels based on Grad-CAM). The bottom row is Grad-CAM masking.}
    \label{fig:PixelFlipping}
\vspace{-8pt}
\end{figure}

\begin{figure}[!t]
\centering
\includegraphics[width = 0.48\textwidth]{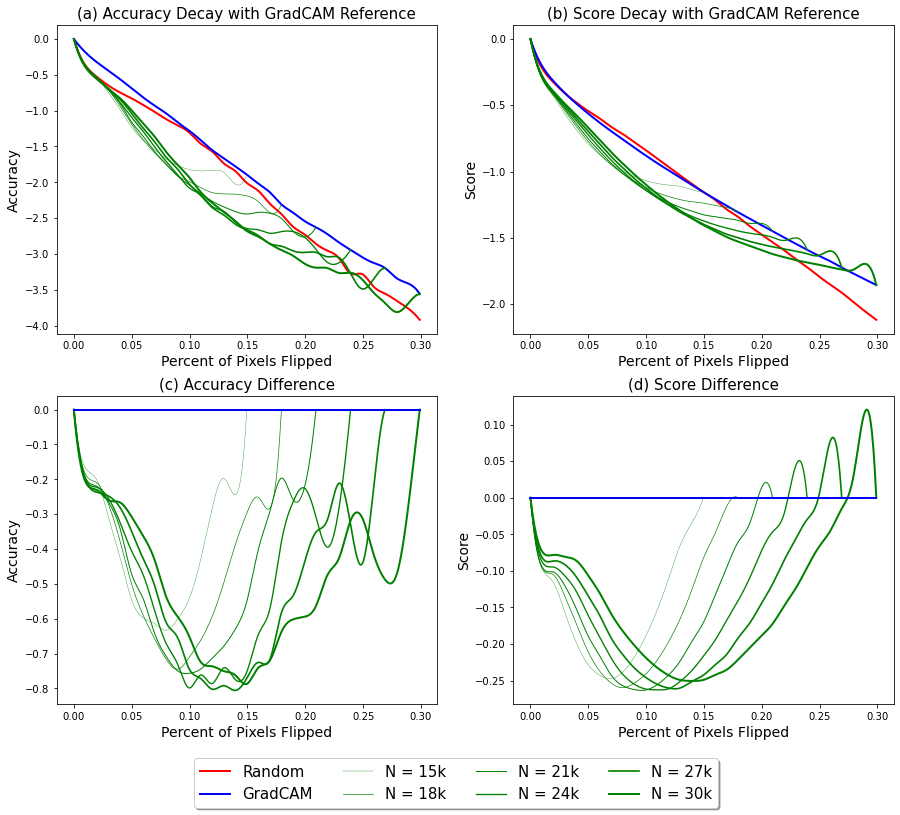}
\caption{Pixel Flipping results with Grad-CAM reference. Blue, red, green curves represent Grad-CAM, Random, and Random with Grad-CAM reference, respectively. Different widths of green curves represent different numbers of referenced pixels $N$. (a)(c) are the results of accuracy and (b)(d) are the results of scores. (c)(d) are the differences between the referenced masking and the Grad-CAM (reference) masking.
All results are normalized to $[0,1]$ and then taken the logarithm for better illustration.}
\label{fig:PixelFlipping_reference}
\vspace{-10pt}
\end{figure}

Now that both the out-of-distribution issue (pixel isolation) and the information of pixels contribute to the performance decay in Pixel Flipping, it is worth exploring how they interact and influence the performance decay. To compare the random masking benchmark with attribution-based masking more fairly, we carry out a Reference Pixel Flipping experiment. The $N$ reference pixels are first selected based on the attribution method, then $n(<N)$ pixels are randomly masked within the $N$ reference pixels. The illustration of such masking is shown in \cref{fig:PixelFlipping}.
{In this way, by confining the random masked area as similar to the Grad-CAM as possible. we minimize the influence caused by information of pixels.} We thus prove the following proposition.

\begin{proposition}\label{prop1}
Let $I^*_N$ be the reference set such that $\forall 0< N\le d$, $|I^*_N| = N$, and $I^*_N\subset I^*_{N+1}$. $\forall 0< n\le N$, let $I_{n,N}\subset I^*_N$ be a random subset of the cardinality $n$. The expected Dice similarity~\cite{dice1945measures} between $I_{n,N}$ and $I^*_n$ is deceasing w.r.t. $N$ in inverse proportion.
\end{proposition}

Please refer to \cref{app:PixelFlipping_reference} for the proof. Let $d\ge N\ge n$ denote the number of all pixels, referenced pixels, and masked pixels, respectively. Based on this proposition, if $N = d$, the referenced masking degenerates to totally random masking. If $N = n$, it degenerates to Grad-CAM masking. So we need to balance $N$ between $n$ and $d$. We let $N$ vary from 15k to 30k, and $n$ vary from 0 to $N$. 
{It should be noticed that this is an approximation, because there are three variables, the masked area (object or background), the pixel isolation, and the number of masked pixels, and it is impossible to alter only one of them without changing others. We fix the number of masked pixels and make compromises between the area and the isolation.} 
The results are shown in \cref{fig:PixelFlipping_reference}. As the combinations of random masking and Grad-CAM masking, all referenced masking results outperform these two ingredients. This corresponds to the fact that isolated masking methods such as Gradient outperform unisolated masking methods like Grad-CAM. Also, from \cref{fig:PixelFlipping_reference}(c)(d), the lowest point of each referenced masking curve locates near the midpoint of the corresponding $N$. This suggests that both isolation and the information of pixels contribute greatly to the performance decay. As $n\rightarrow N$, the effect of isolation diminishes, so the referenced masking converges to the Grad-CAM reference.

\subsubsection{Baseline Flaws}
When a pixel is masked, numerically, there must be some other value to replace it, which is often referred to as the baseline~\cite{sundararajan2017axiomatic}. The baseline represents the null feature in the input that provides no information. It is generally set to simple intuitive values like zeros, means, random noises, etc. Recently, there are other complex baselines such as Gaussian blur~\cite{sturmfels2020visualizing}, or even inpainting algorithms~\cite{samek2021explaining}. Unfortunately, {there is} no baseline theoretically justified as ``truly non-informative". All proposed baselines are approximations and make compromises. 
{Seeking the best baseline is beyond the scope of our work. Therefore, as we want to introduce as little exogenous information as possible, we use simple baselines such as zeros or random noises in experiments.}

{ 
\section{Application of Faithfulness and Plausibility}
\label{sec:suggestion}
Based on the importance and the chaos of plausibility and faithfulness, we present several applications of them with the goal of improving the interpretability research.

\textbf{Stop using plausibility solely as the measure.}
It has been argued that visualizations should not be used solely as the measure for explanations~\cite{leavitt2020towards}. We emphasize that not only visualizations, but also their superset, the plausibility, should not be used solely. Compared with visualizations, it includes quantitative plausibility-based metrics, such as Pointing Game, BAM, etc. Even though plausibility has been extensively applied in XAI, an implausible explanation is not necessarily less desired than a more plausible one. They should not be intentionally avoided. 
In fact, it is those implausible explanations that are really useful. Because they may be actually revealing the implausible mechanism of black boxes.

\textbf{Faithfulness should be handled carefully.}
For DNNs, the ground truth of faithfulness of explanations still remains an open question~\cite{jacovi2020towards}. Currently, with existing techniques, it is impossible to decide directly whether the explanation is faithful or not. We suggest that researchers should not claim faithfulness unless this issue has been solved. {This does not mean faithfulness should be neglected. On the contrary, it should be carefully considered in developing and evaluating explanation methods.}

\textbf{Metrics should introduce as little bias as possible.} With strong assumptions and exogenous information introduced, a metric can be completely biased as shown in \cref{sec:case_studies}. Such evaluation will only lead to counter-productive results.

\textbf{The Goals of explanation metrics and methods should be clear and compatible.} Different explanations methods can serve different goals. We suggest that it is unreasonable to use universal metrics to evaluate these explanations altogether. Therefore, the goal of a metric needs to be clear and consistent with the explanation. Simultaneously, the goals of explanations should also be clear. 


\textbf{It is the usage instead of the superiority of explanations that matters.} Given the insufficient study on the metrics, we encourage researchers not to be stuck in this chaos. Instead of trying to demonstrate the superiority of their explanation methods to others, the usage of explanation methods should draw more attention. An explanation is really desired if it can help in practical applications, such as debugging, or avoiding the overfitting phenomenon.

\vspace{-3pt}
\section{Conclusion}
\label{sec:conclusion}
From the perspective of evaluating explanations, we study existing explanation metrics of two categories, the alignment metrics, and the performance metrics. Based on this, we analyze fatal flaws in nowadays' XAI research from many aspects. We then demonstrate that such flaws are caused by the chaos of explanation metrics. We also present several experiments to show the importance of proper metrics. Finally, we propose suggestions on both developing and evaluating explanations for the XAI research. We encourage researchers to take these criteria into careful consideration. 

We admit that building perfect explanations and developing perfect metrics for explanations are two complementary tasks. Neither of them is fulfilled yet. The goal of this paper is to provide a unified study that clearly analyzes them, and that can be potentially useful for exploring such perfection.
}

\clearpage
\appendix
\setcounter{figure}{0}

\section{Related Work Supplement}
\label{app:related_work}
Based on the stage where the explanations are generated, attribution methods can further be divided as post-hoc methods and self-interpretable methods. Post-hoc methods are already included in \cref{sec:related_work}. They are developed to explain pre-trained models (usually DNNs). However, it is also argued that post-hoc methods are not reliable and hence not trusted by humans, and that self-interpretable models are more desired~\cite{rudin2019stop}. Self-interpretability is acknowledged as the property that the explanations and the predictions are generated at the same stage, and that the explanations are directly involved in the predictions. Within this genre, white-box models such as sparse linear models, decision tree, etc. are usually treated as axiomatically self-interpretable. However, their expressiveness is limited by their low complexity. Therefore, there are also deep models arguably claiming self-interpretability by regularizing the models themselves~\cite{alvarez2018towards,chen2018looks,agarwal2020neural,wang2021self}, etc. But these models are still limited by universality, complexity, expressiveness, etc.

\section{PGD Enhancement}
\label{app:PGD_enhancement}

\subsection{Implementation Details}
The standard PGD attack~\cite{madry2017towards} is defined as follows
\begin{align*}
    \x^{t+1} = \Pi_{\x + S}\big(\x^t + \alpha~\mathrm{sgn}(\nabla_{\x}L(\theta,\x,y))\big)\,,
\end{align*}
where $\alpha$ is the step size, and $S$ is the set of allowed perturbations. $S$ is defined as a box area $[-\epsilon,\epsilon]^d$. The PGD enhancement is then defined as
\begin{align*}
    \x^{t+1} = \Pi_{\x + S}\big(\x^t - \alpha~\mathrm{sgn}(\nabla_{\x}L(\theta,\x,y))\big)\,.
\end{align*}
All PGD enhancement examples shown in this paper are generated under the following settings: $\epsilon = 0.3$, $\alpha = 0.03$, and the number of steps is $30$.

\subsection{PGD Enhancement for Other Post-Hoc Explanations}
Here we present illustration of PGD-enhancement results of different post-hoc explanation methods. Apart from Grad-CAM~\cite{selvaraju2017grad} shown in \cref{sec:faith_ration}, we also present Excitation Backpropagation~\cite{zhang2018top} in \cref{fig:PGD_enhancement_excitationBP}, Linear Approximation in \cref{fig:PGD_enhancement_LinearApprox}, Guided Backpropagation~\cite{springenberg2014striving} in \cref{fig:PGD_enhancement_GuidedBP}, Gradient~\cite{simonyan2013deep} in \cref{fig:PGD_enhancement_Gradient}, and DeConvNet~\cite{zeiler2014visualizing} in \cref{fig:PGD_enhancement_deconvnet}. It can be found that The explanations of enhanced images and the untrained model (middle) are more similar to the explanations of raw images and the untrained model (left), rather than those of raw images and the trained model (right). However, the predicted classification results of the middle columns are the same as the right columns, which are correct. The predicted classification results of the left columns are wrong. This phenomenon is because while the predictions of the middle columns and the right columns are same, their true mechanism (for DNNs, is the collection of weights) are completely different. And this difference in \textit{faithfulness} is captured by most explanation methods. However, with a plausibility-based metric, these explanation methods (for the middle columns) will be measured unfairly.

\subsection{Pixel Isolation}
\begin{figure}[htb!]
\centering
\subfloat{
\includegraphics[width=0.15\textwidth]{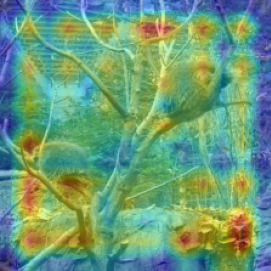}}
\subfloat{
\includegraphics[width=0.15\textwidth]{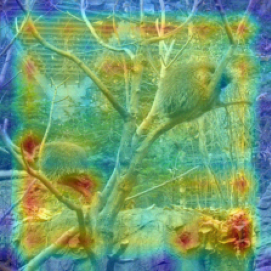}}
\subfloat{
\includegraphics[width=0.15\textwidth]{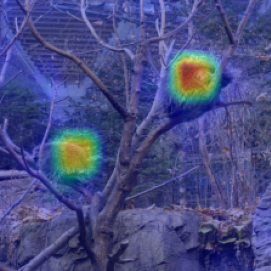}}\\
\subfloat{
\includegraphics[width=0.15\textwidth]{PGD_enhancement/1_ExcitationBP_porcupine_hedgehog_untrainedPredAs_steel-drum.png}}
\subfloat{
\includegraphics[width=0.15\textwidth]{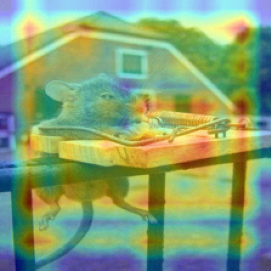}}
\subfloat{
\includegraphics[width=0.15\textwidth]{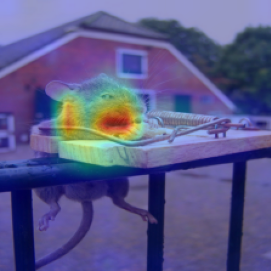}}\\
\subfloat{
\includegraphics[width=0.15\textwidth]{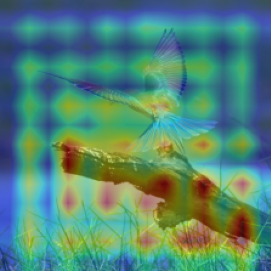}}
\subfloat{
\includegraphics[width=0.15\textwidth]{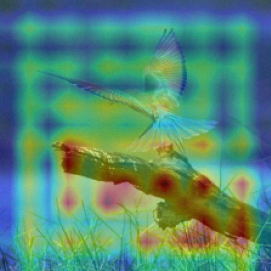}}
\subfloat{
\includegraphics[width=0.15\textwidth]{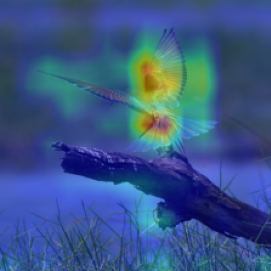}}\\
\setcounter{subfigure}{0}
\subfloat[Untrained]{
\includegraphics[width=0.15\textwidth]{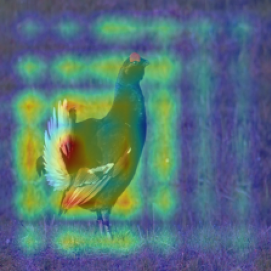}}
\subfloat[Enhanced Untrained]{
\includegraphics[width=0.15\textwidth]{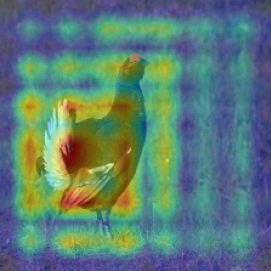}}
\subfloat[Trained]{
\includegraphics[width=0.15\textwidth]{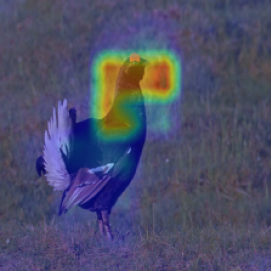}}
\caption{Excitation Backpropagation explanations of VGG-16 models. The results are from the untrained model with raw data as input (top row), the untrained model with PGD-enhanced data as input (middle row), and the trained model with raw data as input (bottom row). All samples of the middle and the bottom rows are correctly classified, and all samples from the top row are falsely classified.}
\label{fig:PGD_enhancement_excitationBP}
\end{figure}

\begin{figure}[htb!]
\centering
\subfloat{
\includegraphics[width=0.15\textwidth]{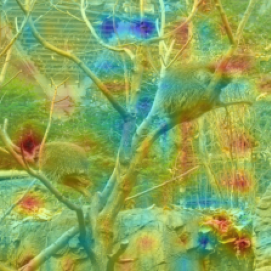}}
\subfloat{
\includegraphics[width=0.15\textwidth]{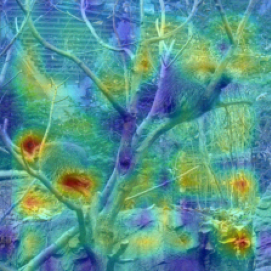}}
\subfloat{
\includegraphics[width=0.15\textwidth]{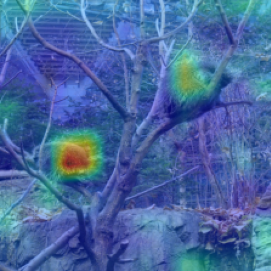}}\\
\subfloat{
\includegraphics[width=0.15\textwidth]{PGD_enhancement/1_LinearApprox_porcupine_hedgehog_untrainedPredAs_steel-drum.png}}
\subfloat{
\includegraphics[width=0.15\textwidth]{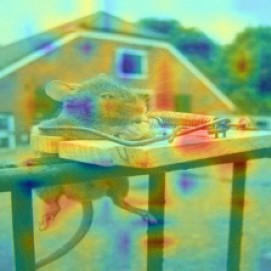}}
\subfloat{
\includegraphics[width=0.15\textwidth]{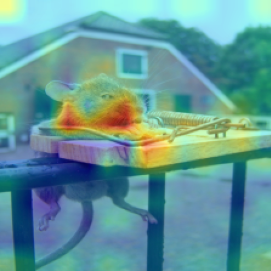}}\\
\subfloat{
\includegraphics[width=0.15\textwidth]{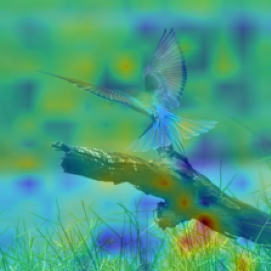}}
\subfloat{
\includegraphics[width=0.15\textwidth]{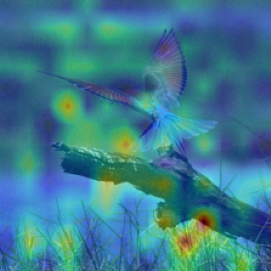}}
\subfloat{
\includegraphics[width=0.15\textwidth]{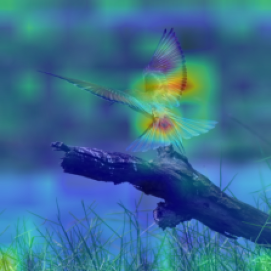}}\\
\setcounter{subfigure}{0}
\subfloat[Untrained]{
\includegraphics[width=0.15\textwidth]{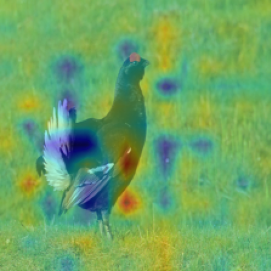}}
\subfloat[Enhanced Untrained]{
\includegraphics[width=0.15\textwidth]{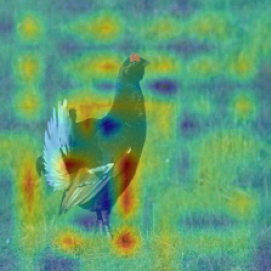}}
\subfloat[Trained]{
\includegraphics[width=0.15\textwidth]{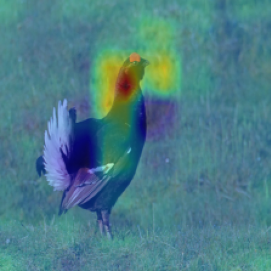}}
\caption
{Linear Approximation explanations of VGG-16 models. The results are from the untrained model with raw data as input (top row), the untrained model with PGD-enhanced data as input (middle row), and the trained model with raw data as input (bottom row). All samples of the middle and the bottom rows are correctly classified, and all samples from the top row are falsely classified.}
\label{fig:PGD_enhancement_LinearApprox}
\end{figure}

\begin{figure}[htb!]
\centering
\subfloat{
\includegraphics[width=0.15\textwidth]{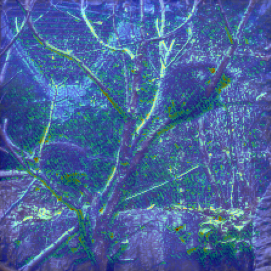}}
\subfloat{
\includegraphics[width=0.15\textwidth]{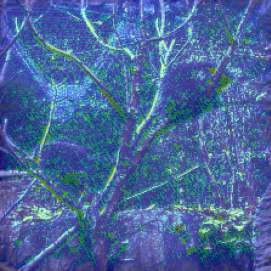}}
\subfloat{
\includegraphics[width=0.15\textwidth]{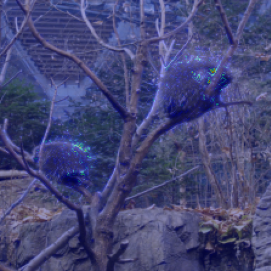}}\\
\subfloat{
\includegraphics[width=0.15\textwidth]{PGD_enhancement/1_GuidedBP_porcupine_hedgehog_untrainedPredAs_steel-drum.png}}
\subfloat{
\includegraphics[width=0.15\textwidth]{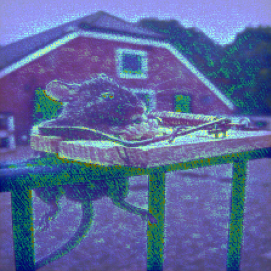}}
\subfloat{
\includegraphics[width=0.15\textwidth]{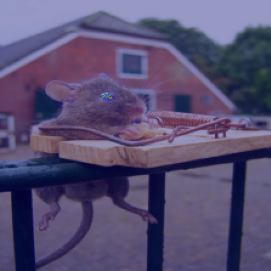}}\\
\subfloat{
\includegraphics[width=0.15\textwidth]{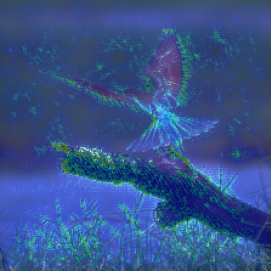}}
\subfloat{
\includegraphics[width=0.15\textwidth]{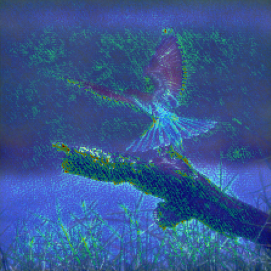}}
\subfloat{
\includegraphics[width=0.15\textwidth]{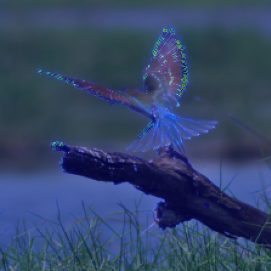}}\\
\setcounter{subfigure}{0}
\subfloat[Untrained]{
\includegraphics[width=0.15\textwidth]{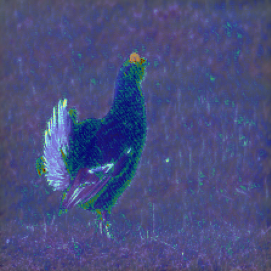}}
\subfloat[Enhanced Untrained]{
\includegraphics[width=0.15\textwidth]{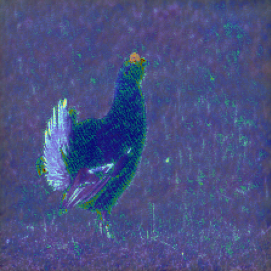}}
\subfloat[Trained]{
\includegraphics[width=0.15\textwidth]{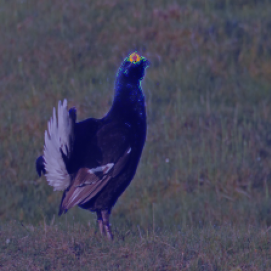}}
\caption
{Guided Backpropagation explanations of VGG-16 models. The results are from the untrained model with raw data as input (top row), the untrained model with PGD-enhanced data as input (middle row), and the trained model with raw data as input (bottom row). All samples of the middle and the bottom rows are correctly classified, and all samples from the top row are falsely classified.}
\label{fig:PGD_enhancement_GuidedBP}
\end{figure}

\begin{figure}[htb!]
\centering
\subfloat{
\includegraphics[width=0.15\textwidth]{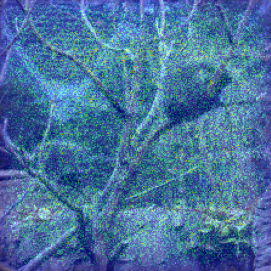}}
\subfloat{
\includegraphics[width=0.15\textwidth]{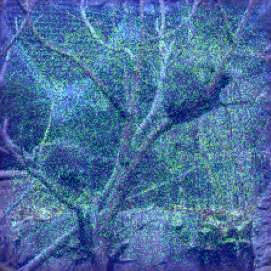}}
\subfloat{
\includegraphics[width=0.15\textwidth]{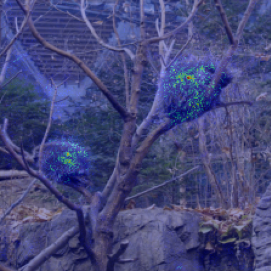}}\\
\subfloat{
\includegraphics[width=0.15\textwidth]{PGD_enhancement/1_Gradient_porcupine_hedgehog_untrainedPredAs_steel-drum.png}}
\subfloat{
\includegraphics[width=0.15\textwidth]{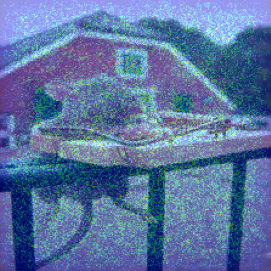}}
\subfloat{
\includegraphics[width=0.15\textwidth]{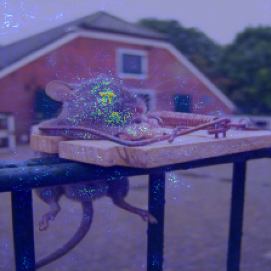}}\\
\subfloat{
\includegraphics[width=0.15\textwidth]{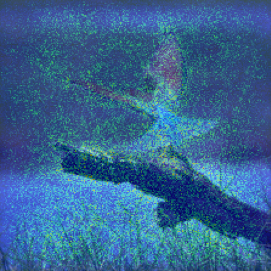}}
\subfloat{
\includegraphics[width=0.15\textwidth]{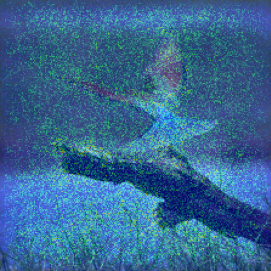}}
\subfloat{
\includegraphics[width=0.15\textwidth]{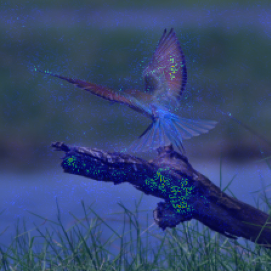}}\\
\setcounter{subfigure}{0}
\subfloat[Untrained]{
\includegraphics[width=0.15\textwidth]{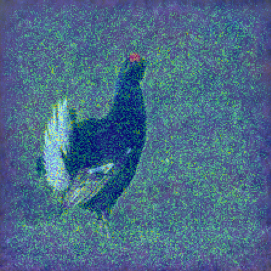}}
\subfloat[Enhanced Untrained]{
\includegraphics[width=0.15\textwidth]{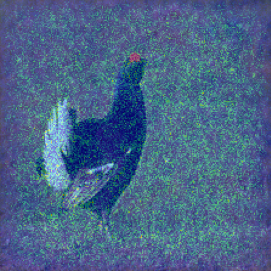}}
\subfloat[Trained]{
\includegraphics[width=0.15\textwidth]{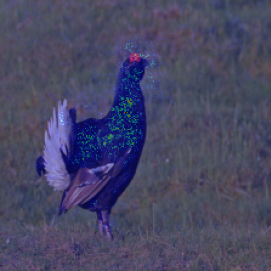}}
\caption
{Gradient explanations of VGG-16 models. The results are from the untrained model with raw data as input (top row), the untrained model with PGD-enhanced data as input (middle row), and the trained model with raw data as input (bottom row). All samples of the middle and the bottom rows are correctly classified, and all samples from the top row are falsely classified.}
\label{fig:PGD_enhancement_Gradient}
\end{figure}

\begin{figure}[htb!]
\centering
\subfloat{
\includegraphics[width=0.15\textwidth]{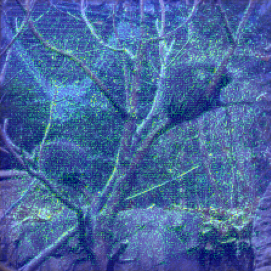}}
\subfloat{
\includegraphics[width=0.15\textwidth]{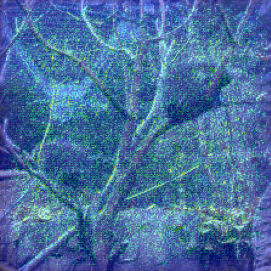}}
\subfloat{
\includegraphics[width=0.15\textwidth]{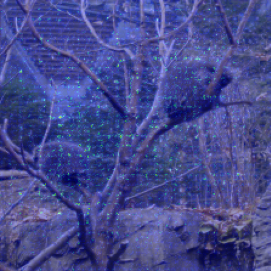}}\\
\subfloat{
\includegraphics[width=0.15\textwidth]{PGD_enhancement/1_DeConvNet_porcupine_hedgehog_untrainedPredAs_steel-drum.png}}
\subfloat{
\includegraphics[width=0.15\textwidth]{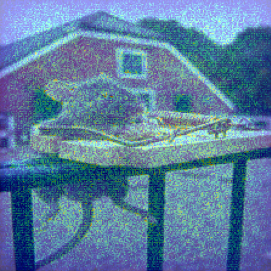}}
\subfloat{
\includegraphics[width=0.15\textwidth]{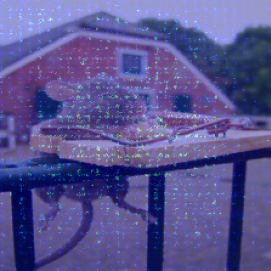}}\\
\subfloat{
\includegraphics[width=0.15\textwidth]{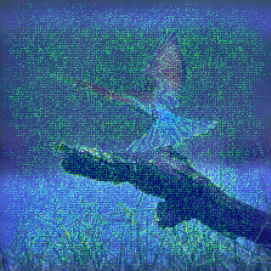}}
\subfloat{
\includegraphics[width=0.15\textwidth]{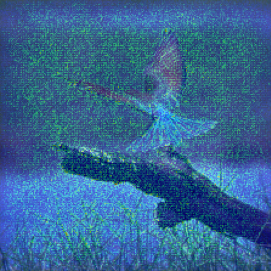}}
\subfloat{
\includegraphics[width=0.15\textwidth]{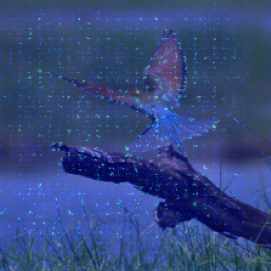}}\\
\setcounter{subfigure}{0}
\subfloat[Untrained]{
\includegraphics[width=0.15\textwidth]{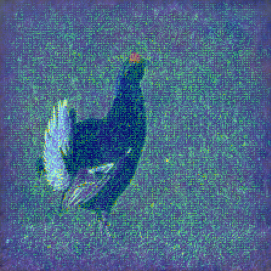}}
\subfloat[Enhanced Untrained]{
\includegraphics[width=0.15\textwidth]{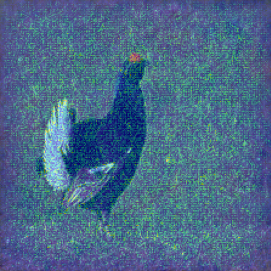}}
\subfloat[Trained]{
\includegraphics[width=0.15\textwidth]{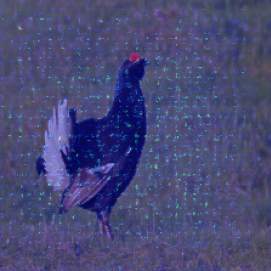}}
\caption
{DeConvNet explanations of VGG-16 models. The results are from the untrained model with raw data as input (top row), the untrained model with PGD-enhanced data as input (middle row), and the trained model with raw data as input (bottom row). All samples of the middle and the bottom rows are correctly classified, and all samples from the top row are falsely classified.}
\label{fig:PGD_enhancement_deconvnet}
\end{figure}

Comparing the explanations of Guided Backpropagation (\cref{fig:PGD_enhancement_GuidedBP}), Gradient (\cref{fig:PGD_enhancement_Gradient}), DeConvNet (\cref{fig:PGD_enhancement_deconvnet}) with others, we can find that the explanations of these three methods have the property \textit{pixel isolation}. That is, the pixels with adjacent attribution values are always isolated. Hence their highlighted areas in heatmaps are many isolated pixels, while the highlighted areas of Grad-CAM (\cref{sec:faith_ration}), Excitation Backpropagation (\cref{fig:PGD_enhancement_excitationBP}), and Linear Approximation (\cref{fig:PGD_enhancement_LinearApprox}) are consecutive pixel patches. This difference is one of the most exogenous information introduced when applying Pixel Flipping or other metrics that requires modifying the input images. More details about the studies on this difference are presented in \cref{sec:case_studies}.

\section{Pointing Game}
\label{app:pointing_game}
Given an image instance $\x$, a classifier $F$, and a targeted class $t$, an attribution method usually returns a attribution map as the explanation for the targeted class. From the quantitative attribution map, there's a pixel that can be recognized as the ``most important''. Pointing Game then calculates the ratio that the most important pixel falls into the ``object areas''. Such areas are usually in the form of pre-annotated bounding boxes or silhouettes of the objects. Formally, suppose there are $N$ instances, and the 2-D coordinate of the most important pixel of $\x_i$ is $(u_i,v_i)$, and the annotated object area of $\x_i$ for target $t_i$ is $\Omega_i$. The explanation method scores a hit if 
$\max_{(u,v)\in\Omega_i}\|(u_i,v_i) - (u,v)\|_2\le \tau$ 
where $\tau$ is a predefined tolerance, usually applied to mitigate the error introduced by upsampling in some attribution methods. Then the Pointing Game score is defined by the ratio of number of total hits to the number of all samples, as
\begin{align*}
    r = \frac{\textrm{\# of hits}}{\textrm{\# of samples}}.
\end{align*}

\section{Annotated MNIST}
\label{app:MNIST}
\begin{figure}[htb!]
    \centering
    \includegraphics[width = 0.45\textwidth]{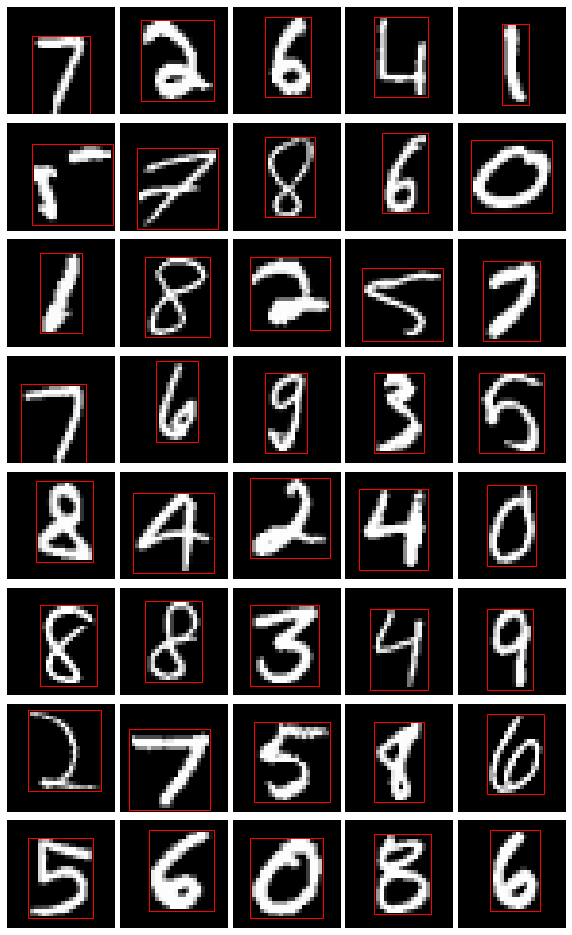}
    \caption{MNIST Data Examples with Bounding Box annotations.}
    \label{fig:MNIST_BBox}
\end{figure}

Since the background pixels of MNSIT data all have value 0, we automatically drawing a bounding box for each image of MNSIT dataset, such that all pixels with positive values are included in the bounding box. The edges of bounding boxes are drawn exactly next to the marginal pixels of the corresponding image margins. Some examples are shown in \cref{fig:MNIST_BBox}. In Pointing Game, an explanation method scores a hit if the pixel with the highest attribution value falls strictly within (not on the edges) the corresponding bounding box. Also, since in the experiments we apply linear models, which do not require upsampling, the tolerance $\tau$ is set to zero.

\section{Accuracy as Indicator in Pixel Flipping}
\label{app:appendix_accuracyPixelFlipping}
Suppose we have a linear classifier, where we treat the weight $\w_i\odot \x$ as the explanations. Suppose we have $\x = [x_1,\cdots,x_d]\in\mathbb{R}^d$, and $W = [\w_1,\cdots,\w_c]\in\mathbb{R}^{d\times c}$, then the prediction is $\y = W^\top \x\in\mathbb{R}^c$. Here we omit the softmax activition since it does not change the relative relations. The classification result is then decided by $\hat{y} = \argmax_{1\le i\le c} y_i$. Suppose $t\in[c]$ to be the target. Since most attribution methods only build the attribution map based on a specific class, (That is, there are $c$ attribution maps, each of which is corresponding to a specific class.) Pixel Flipping method only flips the values of $x_i$ according to the corresponding value $w_t^i$. Without loss of generality, we set $w_t^1x_1\ge w_t^2x_2\ge \cdots w_t^dx_d$ so that we can process the masking by this order. Then by the criterion of Pixel Flipping, we mask $x_1,x_2,\cdots,x_n$ one by one. Suppose the instance $\x$ is classified correctly, then without loss of generality, we can also set $\w_1^\top\x \ge \w_2^\top\x \ge \cdots\ge \w_c^\top\x$. Based on the setting specified above, Pixel Flipping can be easily disproved. The original prediction process is
\begin{align*}
    \begin{bmatrix}
    y_1\\
    y_2\\
    \vdots\\
    y_c
    \end{bmatrix}
    =
    \begin{bmatrix}
    w_1^1 & w_1^2 & \cdots & w_1^d\\
    w_2^1 & w_2^2 & \cdots & w_2^d\\
    \vdots & \vdots & \ddots & \vdots\\
    w_c^1 & w_c^2 & \cdots & w_c^d
    \end{bmatrix}
    \begin{bmatrix}
    x_1\\
    x_2\\
    \vdots\\
    x_d
    \end{bmatrix}
\end{align*}
After masking the most relevant feature, we have

\begin{align*}
    \begin{bmatrix}
    y_1 - w_1^1x_1\\
    y_2 - w_2^1x_1\\
    \vdots\\
    y_c - w_c^1x_1
    \end{bmatrix}
    =
    \begin{bmatrix}
    w_1^1 & w_1^2 & \cdots & w_1^n\\
    w_2^1 & w_2^2 & \cdots & w_2^n\\
    \vdots & \vdots & \ddots & \vdots\\
    w_c^1 & w_c^2 & \cdots & w_c^n
    \end{bmatrix}
    \begin{bmatrix}
    0\\
    x_2\\
    \vdots\\
    x_n
    \end{bmatrix}
\end{align*}
Trivially, although $y_1 = \w_1^\top\x \ge \w_2^\top\x = y_2$, and $w_1^1x_1 > w_1^ix_i$ for $\forall i>1$, it is not guaranteed that masking $x_1$ can result in the best performance decay. This is because the difference between the distributions of different weight vector $\w_i$. This can be verified by a very simple example. Let
\begin{align*}
    \x = 
    \begin{bmatrix}
    1\\
    1
    \end{bmatrix}
    ,W = 
    \begin{bmatrix}
    \w_1 & \w_2 & \w_3
    \end{bmatrix}
    =
    \begin{bmatrix}
    2 & \sqrt{6} & \sqrt{5}\\
    1 & 0 & 0
    \end{bmatrix}
\end{align*}
We then have the preconditions $y_1 = 3 > \sqrt{6} = y_2 > \sqrt{5} = y_3, w_1^1x_1 = 2 > 1 = w_1^2x_2$. However, masking $x_1$ will result in 
\begin{align*}
    y_1' = y_1 - w_1^1x_1 = 1\\
    y_2' = y_2 - w_2^1x_1 = 0\\
    y_3' = y_3 - w_3^1x_1 = 0
\end{align*}
The classification result will not be changed. If we instead mask the feature $x_2$, which has lower explanation value, we then have
\begin{align*}
    y_1' =& y_1 - w_1^2x_2 = 2\\
    y_2' =& y_2 - w_2^2x_2 =\sqrt{6}\\
    y_3' =& y_3 - w_3^2x_2 =\sqrt{5}  
\end{align*}
which changes the accuracy. This example shows that accuracy should not be used as the indicator, even for the linear model.

\section{Proof of \cref{{prop1}}}
\label{app:PixelFlipping_reference}
Suppose $\x = (x_1,\cdots, x_d)$ is the input data. Without loss of generality, suppose they are ordered monotonically decreasing by the order of attribution values from some explanations. $I^*_n = [n]$, and $I_{n,N}\subset [N]$ are two index subsets, where $I_{n,N}$ is randomly sampled such that $|I_{n,N}| = n$. Denote the Dice similarity as $s(A, B) = \frac{2|A\cap B|}{|A| + |B|}$, then the expected distance between $I_{n,N}$ and $I^*_n$ is defined as
\begin{align*}
    D(N) =& \ep_{I_{n,N}\subset I^*_N}[s(I_n^*,I_{n,N})]\\
    =& \sum_{k=\max\{0, 2n-N\}}^n\frac{\binom{n}{k}\binom{N-n}{n-k}}{\binom{N}{n}}\frac{k}{n}\\
    =& \frac{1}{n\binom{N}{n}}\sum_{k=\max\{0, 2n-N\}}^nk\binom{n}{k}\binom{N-n}{n-k}\\
    =& \frac{1}{n\binom{N}{n}}\cdot n\binom{N-1}{n-1}
\end{align*}

which is a inverse proportional function. The above derivation is based on the following reasoning. According to the binomial theorem, on the one hand
\begin{align*}
    n(1+x)^{N-1} = \sum_{k=1}^{N}n\binom{N-1}{k-1}x^{k-1}
\end{align*}
on the other hand,
\begin{align*}
    n(1+x)^{N-1} =& n(1+x)^{n-1}(1+x)^{N-n}\\
    =& n\Big(\sum_{i=0}^{n-1}\binom{n-1}{i}x^i\Big)\Big(\sum_{j=0}^{N-n}\binom{N-n}{j}x^j\Big)\\
    \overset{_{k=i+j}}{=\joinrel=\joinrel=}& \sum_{i=1}^n\sum_{k=i}^{N-n+i}i\binom{n}{i}\binom{N-n}{k-i}x^{k-1}\\
    =& \sum_{k=1}^{N-n}\sum_{i=\max\{1, k-N+n\}}^{\min\{k,n\}}i\binom{n}{i}\binom{N-n}{k-i}x^{k-1}
\end{align*}
Comparing the coefficients of $x^{n-1}$, we have
\begin{align*}
    \sum_{i=\max\{0, 2n-N\}}^ni\binom{n}{i}\binom{N-n}{n-i} = n\binom{N-1}{n-1}.
\end{align*}

Therefore, $D(N) = \frac{1}{n\binom{N}{n}}\cdot n\binom{N-1}{n-1} = \frac{n}{N}.\qedsymbol$

\end{document}